\documentclass[10pt,twocolumn,letterpaper]{article}

\usepackage{cvpr}              

\usepackage{graphicx}
\usepackage{amsmath}
\usepackage{amssymb}
\usepackage{booktabs}
\usepackage{multirow}
\usepackage{bbding}
\usepackage{ulem}

\usepackage{threeparttable}

\usepackage[pagebackref,breaklinks,colorlinks]{hyperref}

\usepackage[capitalize]{cleveref}
\crefname{section}{Sec.}{Secs.}
\Crefname{section}{Section}{Sections}
\Crefname{table}{Table}{Tables}
\crefname{table}{Tab.}{Tabs.}

\def\cvprPaperID{4562} 
\def\confName{CVPR}
\def\confYear{2022}

\begin{document}


\title{ViSTA: Vision and Scene Text Aggregation for Cross-Modal Retrieval}

\makeatletter
\newcommand{\printfnsymbol}[1]{%
  \textsuperscript{\@fnsymbol{#1}}%
}
\makeatother

\author{Mengjun Cheng$^{2}$\thanks{\noindent Equal Contributions. This work is done when Mengjun Cheng is a research intern at Baidu Inc.} \quad Yipeng Sun$^1$\printfnsymbol{1}\thanks{Corresponding author.} \quad Longchao Wang$^1$ \quad Xiongwei Zhu$^1$ \\
\quad Kun Yao$^1$ \quad Jie Chen$^2$ \quad Guoli Song$^3$ \quad Junyu Han$^1$ \quad Jingtuo Liu$^1$ \quad Errui Ding$^1$ \quad Jingdong Wang$^1$\\
Department of Computer Vision Technology (VIS), Baidu Inc.$^1$ \\
\quad Department of Computer Science, Peking University$^2$ \quad Peng Cheng Laboratory$^3$\\
{\tt\small \{sunyipeng, wanglongchao, zhuxiongwei, yaokun01, hanjunyu, liujingtuo\}@baidu.com}\\
{\tt\small \{chengmengjun, chenjie\}@pku.edu.cn, \tt\small \{dingerrui, wangjingdong\}@baidu.com}}
\maketitle

\begin{abstract}
Visual appearance is considered to be the most important cue to understand images for cross-modal retrieval, while sometimes the scene text appearing in images can provide valuable information to understand the visual semantics. Most of existing cross-modal retrieval approaches ignore the usage of scene text information and directly adding this information may lead to performance degradation in scene text free scenarios. 
To address this issue, we propose a full transformer architecture to unify these cross-modal retrieval scenarios in a single \textbf{Vi}sion and \textbf{S}cene \textbf{T}ext \textbf{A}ggregation framework (ViSTA). Specifically, ViSTA utilizes transformer blocks to directly encode image patches and fuse scene text embedding to learn an aggregated visual representation for cross-modal retrieval. To tackle the modality missing problem of scene text, we propose a novel fusion token based transformer aggregation approach to exchange the necessary scene text information only through the fusion token and concentrate on the most important features in each modality. To further strengthen the visual modality, we develop dual contrastive learning losses to embed both image-text pairs and fusion-text pairs into a common cross-modal space. 
Compared to existing methods, ViSTA enables to aggregate relevant scene text semantics with visual appearance, and hence improve results under both scene text free and scene text aware scenarios. Experimental results show that ViSTA outperforms other methods by at least $\bf{8.4}\%$ at Recall@1 for scene text aware retrieval task. Compared with state-of-the-art scene text free retrieval methods, ViSTA can achieve better accuracy on Flicker30K and MSCOCO while running at least three times faster during the inference stage, which validates the effectiveness of the proposed framework.
\end{abstract}

\begin{figure}[ht]
   \includegraphics[width=1\columnwidth]{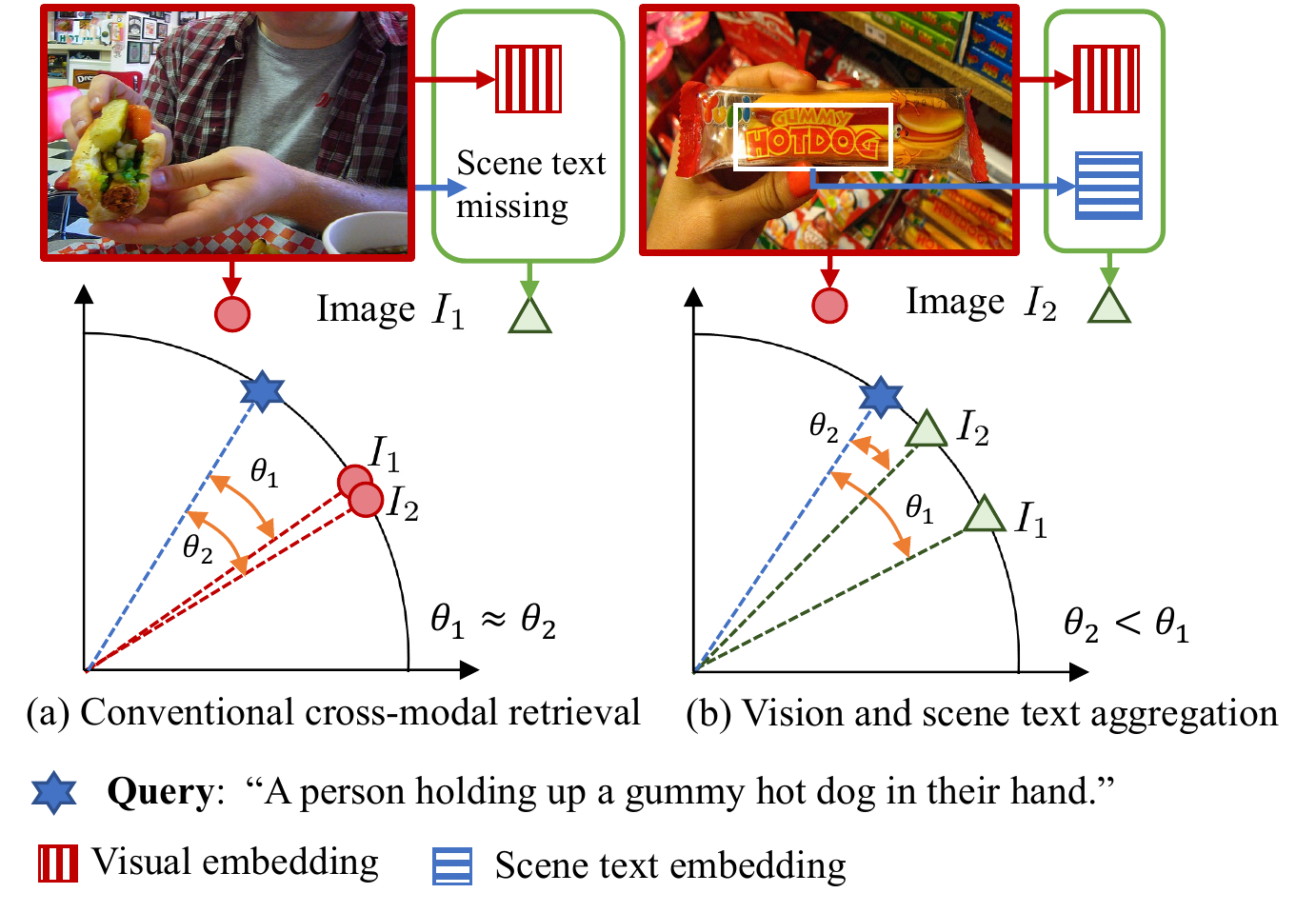}\vspace{-1em}
   \caption{Given a text query, two images are close in visual semantics for (a) conventional cross-modal retrieval. By considering visual appearance and scene text information, e.g.,``gummy hotdog", into one framework, (b) the proposed \textbf{Vi}sion and \textbf{S}cene \textbf{T}ext \textbf{A}ggregation (ViSTA) approach enables to distinguish the semantic difference between images $I_{1}$ and $I_{2}$ ($\theta_{2} < \theta_{1}$), and can be also adapted to conventional scene text free scenarios.}\vspace{-1.25em}
   \label{fig:motivation}
\end{figure}

\vspace{-1em}
\section{Introduction}\label{sec:intro}
As one of the most important multi-modal understanding tasks, cross-modal retrieval attracts much attention due to its valuable applications, e.g., news search and product retrieval. Cross-modal text-to-image retrieval~\cite{frome2013devise,faghri2017vse++,SCAN} aims to return the most relevant candidate based on the relevance between the text content of a query and the visual appearance of an image. The performance of this retrieval task is largely improved by better 
visual representation and detailed image-text alignment~\cite{SCAN, VSRN, GSMN, IMRAM}. 

\vspace{-0.08em}
In recent years, following the success of BERT~\cite{BERT} in natural language modeling, transformer-based single encoder architectures~\cite{vilbert, imagebert, unicoder, UNITER, oscar, ERNIE-ViL, vinvl, Pixel-BERT, SOHO, ViLT, xue2021probing} are adopted to fuse images and text, and image-text pre-training for fine-tuning becomes the mainstream paradigm in modeling visual-language tasks, significantly boosting the cross-modal retrieval performance. However, these approaches with deep interactions between images and text are orders of magnitudes slower and hence impractical for the large-scale cross-modal retrieval task. As dual-encoder architectures, CLIP~\cite{CLIP}, ALIGN~\cite{ALIGN} and WenLan~\cite{wenlan} exploit cross-modal contrastive pre-training by encoding images and text separately, which allows that image and text features can be computed in an offline setting to efficiently calculate similarities between large-scale image-text pairs. Even though the performance of the cross-modal retrieval task is greatly improved by the million-scale image-text contrastive pre-training~\cite{CLIP}, it is still difficult and ineffective to learn specific fine-grained visual concepts, e.g., the scene text semantics from images~\cite{CLIP}. 
More recently, a new cross-modal retrieval task~\cite{STARNet} is proposed to enable the usage of scene text in an image together with its visual appearance. Specifically, an image in this task is paired with the corresponding scene text features to help to determine the similarity between the query's textual content and the image's visual appearance plus scene text. Benefiting from exploiting additional scene text features, this model can improve the cross-modal retrieval accuracy than those exploiting only visual appearance. Nevertheless, in a real-world image corpus, there are only a fraction of images containing scene text instances. 
The model designed for the scene text aware retrieval task might fail to generate reliable similarities between the query and images without scene text instances, and can not be adapted to the conventional scene text free retrieval task.

\vspace{-0.23em} 
To overcome this issue, we propose an effective \textbf{Vi}sion and \textbf{S}cene \textbf{T}ext \textbf{A}ggregation (ViSTA) framework to tackle both scene text aware and scene text free cross-modal retrieval tasks. Specifically, ViSTA utilizes a full transformer design to directly encode image patches and fuse scene text embedding to learn an aggregated visual representation. To enforce each modality focusing on its most important features, we propose a novel token based aggregation approach by sharing the necessary scene text information only through the fusion token. To tackle the modality missing problem of scene text, we further develop dual contrastive supervisions to strengthen the visual modality, and embed both image-text pairs and fusion-text pairs into a common cross-modal space. Compared to existing fusion methods, ViSTA enables to aggregate relevant scene text semantics with visual appearance, and hence improve results under both scene text free and scene text aware scenarios. 

\vspace{-0.23em}
The contributions of this paper are three-fold. $\bf{1)}$ We propose a full transformer architecture to effectively aggregate vision and scene text, which is applicable in both scene text aware and scene text free retrieval scenarios. $\bf{2)}$ We propose a fusion token based transformer aggregation design to exchange the relevant information among visual and scene text features, and dual contrastive losses to enhance visual features. $\bf{3)}$ The proposed cross-modal retrieval framework can remarkably surpass existing methods for the scene text aware retrieval task and achieve better performance than state-of-the-art approaches on scene text free retrieval benchmarks as well.

To the best of our knowledge, it is the first time to solve scene text free and scene text aware cross-modal retrieval tasks with a vision and scene text aggregated transformer.


\begin{figure*}[ht]
\begin{center}
   \includegraphics[width=1.9\columnwidth]{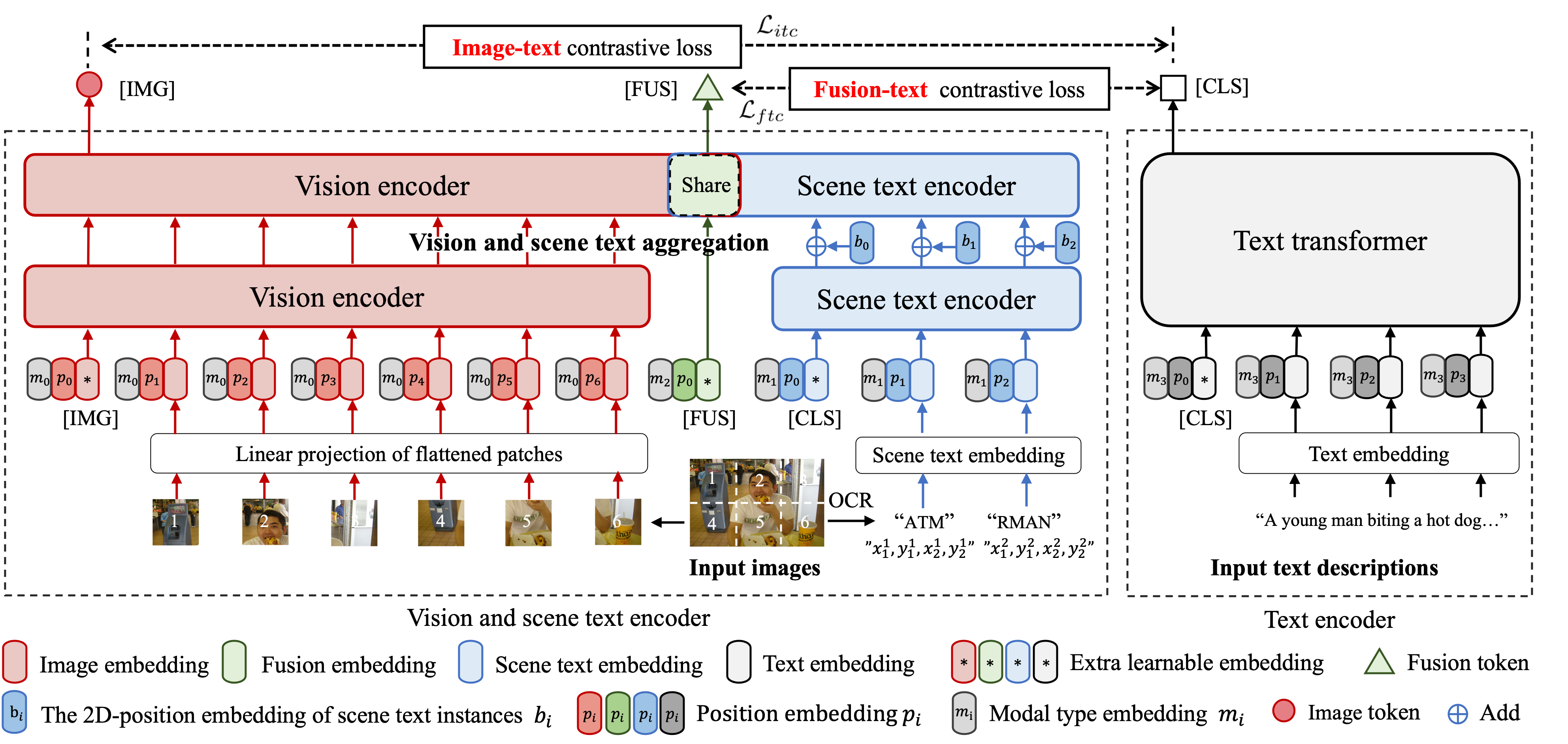}
\scriptsize   
    \end{center}\vspace{-2em}
\scriptsize
   \caption{The proposed \textbf{Vi}sion and \textbf{S}cene \textbf{T}ext \textbf{A}ggregation~(ViSTA) framework for cross-modal retrieval. With the proposed fusion token based vision scene text aggregation layer, ViSTA learns a common cross-modal space by a dual-encoder transformer architecture, supervised by dual contrastive losses between image-text pairs and fusion-text pairs, respectively.}
   \label{fig:framework}
\end{figure*}

\section{Related Work}
\noindent \textbf{Cross-modal retrieval}
aims to return relevant images or text descriptions given text or an image query. Most approaches learn a joint cross-modal embedding space to produce closer representation for semantically relevant image and text pairs~\cite{Skip-Gram, frome2013devise, faghri2017vse++}. Since the deep learning era, the visual representation for cross-modal retrieval has been consistently improved from grid-based CNN~(convolution neural network)~\cite{faghri2017vse++} to a pre-trained object detector~\cite{SCAN, VSRN}. In the meantime, finer image-text alignment approaches are developed, e.g., attention mechanisms, iterative matching, and graph-based relationship reasoning between image features and text embedding~\cite{SCAN, VSRN, IMRAM, GSMN, SGRAF}. Most of these approaches rely on RoI~(region-of-interest) features extracted from a pre-trained Faster-RCNN detector on the Visual Genome~(VG) dataset~\cite{VG_dataset}, which limits the performance on the out-of-domain visual concepts. By contrast, ViSTA directly takes image patches as the input and builds upon the recent contrastive image-text pre-training paradigm, which is capable of achieving better performance by end-to-end training at a much faster inference speed.

\noindent \textbf{Vision language pre-training}
has become a mainstream paradigm in multi-modal understanding, which can remarkably boost the performance on various vision and language tasks, e.g., cross-modal retrieval and visual question answering~(VQA), etc. Most of these approaches utilize transformer based architectures, which can be categorized as single-encoder and dual-encoder pre-training. The single encoder architectures~\cite{vilbert, vl-betr, imagebert, tan2019lxmert, unicoder, UNITER, oscar, ERNIE-ViL, vinvl, Pixel-BERT, SOHO, ViLT, xue2021probing} are adopted to fuse images and text with the multi-modal transformer for interactions, performing high accuracy in various downstream tasks. To speed up the inference stage and adapt to more visual categories, grid-based image features~\cite{Pixel-BERT, SOHO} and newly proposed patch-based image embedding methods~\cite{ViLT, xue2021probing, albef} are utilized for end-to-end training, which directly take image pixels or patches and text embedding as the input. However, the computation cost of these approaches is still huge and impractical for the large-scale cross-modal retrieval task. Instead, dual-encoder architectures~\cite{CLIP,ALIGN,wenlan} encode images and text separately, making it possible to calculate similarities of image-text pairs in the linear time complexity. Even though the performance of the cross-modal retrieval task is greatly improved by the million-scale image-text contrastive pre-training~\cite{CLIP}, it is still difficult and ineffective to learn specific fine-grained visual concepts, e.g., the scene text semantics from images~\cite{CLIP}. By contrast, ViSTA incorporates vision and scene text into a full transformer based dual-encoder architecture, taking image patches, scene text, and text queries as the input for unified cross-modal retrieval.

\noindent \textbf{Scene text in vision and language}
receives much attention as the extension of previous applications, e.g., text-based image caption \cite{sidorov2020textcaps,yang2021tap} and Text-VQA \cite{singh2019towards, biten2019scene,yang2021tap,Beyond_OCR_VQA,simple_is_not_easy}. All these approaches utilize OCR~(optical character recognition) results to form scene text embedding~\cite{singh2019towards, biten2019scene,yang2021tap,single_shot_st}, following the typical architecture of single-stream transformer~\cite{vilbert} with RoI region features. Other works \cite{single_shot_st}\cite{wang2021scene_CVPR21} for scene text retrieval tasks aim to return images that contain the query word, and a CNN based fusion approach~\cite{integrate_scene_text} integrates scene text and visual appearance to improve the performance for fine-grained image classification in specific scenarios. More recently, StacMR \cite{STARNet} introduces scene text aware cross-modal retrieval~(StacMR) considering scene text as an additional modality, which utilizes GCN~(graph convolution network) to obtain context representation of images and scene text for final fusion. Different from all these methods, ViSTA utilizes full transformer blocks to encode image patches and scene text with mid-level fusion, which can be adapted to both scene text aware and scene text free scenarios.

\section{Approach}
The overall architecture of our proposed ViSTA framework is developed as a dual-encoder architecture as shown in Fig.~\ref{fig:framework}, which makes it practical for large-scale cross-modal retrieval. To achieve a strong feature representation for better retrieval accuracy, we adopt a full transformer design to encode images, scene text, and text query by uni-modal encoders, respectively, before feeding them for further aggregation and calculating the cross-modal contrastive losses. The whole model including vision, scene text, and text encoders is end-to-end trainable, which allows better generalization beyond RoI features by cross-modal pre-training~\cite{Pixel-BERT}\cite{SOHO}\cite{ViLT}\cite{xue2021probing}. In order to fuse visual features with relevant scene text semantics, we propose a fusion token based aggregation approach, which shares the relevant information across these two modalities only through the fusion token. As a result, this token can see all the information at each transformer layer and can be used for fusion-text contrastive learning. Since scene text instances do not often appear in images and in some cases the correlation between scene text and images might be weak in visual semantics. Therefore, to enhance the visual representation rather than over-fit to the noisy scene text features, we also utilize image token at the last layer for effective image-text contrastive learning. With such designs, ViSTA can be effectively adapted to both scene text aware and scene text free retrieval scenarios.

\noindent \textbf{Problem formulation.} Given a set of image and text pairs, the vision and scene text encoder aims to encode an image $I$ and recognize the scene text appearing in this image. The scene text instances contain a set of $N_o$ detected words and locations by an OCR model as $\mathcal{O} = \{\mathbf{o}^{word}_j,\mathbf{o}^{bbox}_j\}_{j=1}^{N_o}$. If there is no scene text detected in the image, $\mathcal{O}$ can be an empty set as $\emptyset$. 
In the scene text aware text-to-image retrieval task~\cite{STARNet}, the model is required to generate a similarity score  $S(q,I)$ between a text query $q$ and each image $I$ based on the relevance of the query's textual content and the image's visual features $\mathcal{V}$ together with its scene text features $\mathcal{O}$. In the scene text free text-to-image retrieval task, which is the same as the conventional text-to-image retrieval, scene text instances do not appear in images. Therefore, these images are only sorted by the relevance between the visual appearance and the content of text query.

\noindent\subsection{Vision and Scene Text Encoders}
\label{Vision Encoder}
Following the success of vision transformer~\cite{ViT}, the vision encoder directly takes image patches as the input. By slicing an image into multiple patches, a patch sequence $\mathbf{P} = [\mathbf{p}_1,\cdots, \mathbf{p}_{N_p}]$ is used to form a simple linear projection of pixels before  feeding into transformers. A positional embedding is added to each patch token to encode the position information. Besides, the embedding of the devised special token \textup{[IMG]} is inserted in $\mathbf{P}$. The vision encoder is built upon a stack of $L_v$ standard transformer layers. Let us denote the input sequence of the $l$-th vision transformer layer by $\mathbf{V}_l$. The output sequence of the $l$-th layer severs as the input sequence of the next layer, calculated as
\begin{equation}
\begin{split}
\label{equ_workflow}\vspace{-0.5em}
    &\mathbf{Y}_l \gets  \mathrm{MHSA}(\mathrm{LN}(\mathbf{V}_l)) + \mathbf{V}_l\\
     &\mathbf{V}_{l+1} \gets \mathrm{MLP}(\mathrm{LN}(\mathbf{Y}_l)) + \mathbf{Y}_l,\\
\end{split}
\end{equation}
where $\mathrm{MHSA}(\cdot)$ denotes the multi-head self-attention layer,  $\mathrm{MLP}(\cdot)$ denotes the multi-layer perception layer, and $\mathrm{LN}(\cdot)$ denotes the layer normalization. The input of the first transformer block, $\mathbf{V}_1$, is just the patch sequence $\mathbf{P}$.  Finally, the output of the last vision transformer layer, $\mathbf{V}_{L_v}$, serves as the visual features $\mathcal{V} = \{\mathbf{v}_j\}_{j=1}^{N_v}$. 
To be specific, the $j$-th item in $\mathbf{V}_{L_v}$ corresponds to the $\mathbf{v}_j$ as $ \mathbf{v}_j = \mathbf{V}_{L_v}[:,j]$.




Similar to the vision encoder, the scene text encoder is a stack of $L_s$ standard transformer layers. The input scene text embedding is mainly obtained from the OCR results by Google API \cite{Google_api} and encoded in tokens. The input token from these OCR results is combined with modal type $\mathbf{S}^{type}$ and position embedding $\mathbf{S}^{token\_id}$ as
\begin{equation}
    \mathbf{S}_{init} = \mathrm{Embedding}(\mathbf{o}^{word}) + \mathbf{S}^{type} + \mathbf{S}^{token\_id}. \\
\end{equation}
Following the previous method in Text-VQA~\cite{M4C}, the scene text embedding encoded by BERT \cite{BERT} can be further combined with the $4$-dimensional location information of OCR tokens using normalized bounding box coordinates $\mathbf{o}^{bbox}$ and can be formulated 
as\vspace{-0.5em}
\begin{equation}
\begin{aligned}
    \mathbf{S}_0 &= \mathbf{BERT}(\mathbf{S}_{init}) + \mathrm{F}_{linear}(\mathbf{o}^{bbox}), \\ 
\end{aligned}
\end{equation}
where $\mathrm{F}_{linear}$ linearly projects the normalized coordinates into the two-dimensional position embedding with the same size as the encoded scene text tokens.

\begin{table*}
\caption{Dataset split for the evaluation of cross-modal retrieval tasks. Note that $\ast$ indicates that the CTC-5K test samples have been excluded from the MSCOCO train split.}\vspace{-0.5em}
\label{Tab-setting}
\centering
\setlength{\tabcolsep}{1.2mm}{
\small
\begin{tabular}{c|c|c|c} 
\hline
Task & Pre-training  & Fine-tuning  & Test  \\ 
\hline
Scene text aware & VG  & Flickr30K + TC + CTC train & CTC-1K, 5K  \\ 
\hline
\multirow{2}{*}{Conventional scene text free} & \multirow{2}{*}{SBU + GCC + VG + MSCOCO$^\ast$} & Flickr30K train  & Flickr30K test   \\ 
\cline{3-4} & & MSCOCO$^\ast$ train & MSCOCO-5K test~  \\
\hline
\end{tabular}}
\end{table*}

\begin{figure}[ht]
    \centering
    \includegraphics[width=1.\columnwidth]{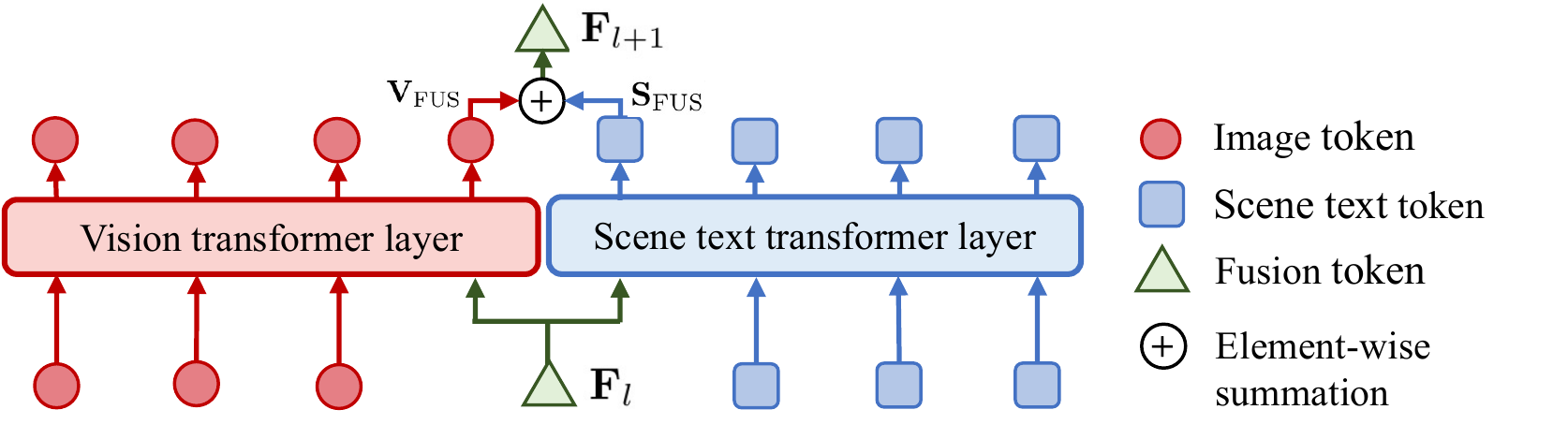}\vspace{-0.5em}
    \caption{The vision scene text aggregation layer. The fusion token shared between two modalities exchanges the relevant information to learn a scene text aggregated visual representation.}\vspace{-1em}
    \label{fig:pipeline}
\end{figure}

\subsection{Vision and Scene Text Aggregation} 
\label{Vision and Scene Text Aggregation}
Since scene text appearing in images may provide valuable information while in most cases images do not contain any scene text information, the semantic relevance between scene text and visual appearance varies case by case, which might be weak in correlations. Therefore, it is challenging to aggregate these two different modalities into a unified visual representation for effective cross-modal retrieval.

To handle both scene text aware and scene text free cross-modal retrieval tasks, it is necessary for the visual tower to learn corresponding final features of image modality for matching. Therefore we use different tokens, an image token or a fusion token, to get the final features based on whether the OCR recognition result is none or not in the training stage.
In the scene text free scenarios, our visual tower degenerates to a pure vision encoder model as in Section \ref{Vision Encoder} and outputs the image feature of [IMG] token as final features.
In the scene text aware scenarios, we use the scene text encoder to learn semantic features of scene text.
And as shown in Fig \ref{fig:framework}, our visual tower simply adds $L_f$ layers of vision and scene text aggregation layer to make mid-level fusion in image modality and outputs fusion features from extra fusion token [FUS] as final features.

As shown in the detailed structure of Fig.~\ref{fig:pipeline}, the vision scene text aggregation layer is composed of a vision transformer layer and a scene text transformer layer from two encoders. 
To exchange the relevant information of vision and scene text, the two layers are added with a new token, which is a shared special fusion token [FUS]. We denote the input image token and scene text token of $l$-th vision encoder and scene text encoder in the aggregation stage by $\mathbf{V}_l$ and $\mathbf{S}_l$. And the input fusion token of $l$-th vision and scene text aggregation is denoted by $\mathbf{F}_l$.
The workflow of the vision transformer layer of Eq. \ref{equ_workflow} in the aggregation stage is updated to
\begin{equation}
\begin{aligned}
    &\mathbf{Y}_l \gets \mathrm{MHSA}(\mathrm{LN}([\mathbf{V}_l; \mathbf{F}_l])) + \left[  \mathbf{V}_l; \mathbf{F}_l \right]\\
    &\left[  \mathbf{V}_{l+1}; \mathbf{V}_{\mathrm{FUS}} \right] \gets  \mathrm{MLP}( \mathrm{LN}(\mathbf{Y}_l)) + \mathbf{Y}_l, \\
\end{aligned}
\end{equation}
where $\mathbf{V}_{\mathrm{FUS}}$ is the output image feature corresponding to the fusion token.
Same is the workflow of the scene text transformer layer in the aggregation stage as
\begin{equation}
\begin{aligned}
    &\mathbf{Y}_l \gets \mathrm{MHSA}( \mathrm{LN}( [  \mathbf{S}_l; \mathbf{F}_l ] )) + \left[  \mathbf{S}_l; \mathbf{F}_l \right]\\
    &\left[  \mathbf{S}_{l+1}; \mathbf{S}_{\mathrm{FUS}} \right] \gets  \mathrm{MLP}(\mathrm{LN}( \mathbf{Y}_l)) + \mathbf{Y}_l, \\
\end{aligned}
\end{equation}
where $\mathbf{S}_{\mathrm{FUS}}$ is the output scene text feature corresponding to the fusion token.
The input fusion features of the next layer are calculated by their element-wise summation as shown in Fig.~\ref{fig:pipeline}, defined as $\mathbf{F}_{l+1} = \mathbf{V}_{\mathrm{FUS}} + \mathbf{S}_{\mathrm{FUS}}$. In this way, visual features $\mathbf{V}$ and scene text features $\mathbf{S}$ are learned through independent transformer layers, respectively. 
The special fusion token [FUS] plays the role of the bridge of two encoders as it is shared in two encoders. 
Due to the vision and scene text aggregation layers, the learning of image features and scene text features is affected by each other by the indirect fusion token. A similar bottleneck attention structure for video classification \cite{bottleneck} fuses video patches and sound by averaging the prediction of the two modalities. Instead of updating shared tokens twice, ViSTA directly adds the predicted fused tokens from vision and scene text transformer layers, forming the fusion token during the aggregation process. To further consider the modality missing problem of scene text, we propose additional image-text contrastive loss to enhance the visual representation together with the fusion-text contrastive loss.
Therefore, both image-text pairs and fusion-text pairs contain the information of visual appearance and share only the relevant part of the information within scene text through shared tokens, which aims to benefit scene text aware cross-modal learning.

\noindent \textbf{Fusion feature embedding.} 
We consider the fusion token as another modality and therefore add a different modal type embedding to the randomly initialized [FUS] token embedding, which can be calculated as
\begin{equation}
\begin{aligned}
\mathbf{F}_0 = \mathbf{F}^{init} + \mathbf{F}^{type} + \mathbf{F}^{token\_id},
\end{aligned}
\end{equation}
where $\mathbf{F}_0$ is the first input fusion features of vision and scene text aggregation layers.


\begin{table}
\centering
\caption{Model settings at various scales.}\vspace{-0.5em}
\label{Tab-setting-model}
\small
\setlength{\tabcolsep}{1.8mm}{
\begin{tabular}{c|cccc} 
\hline
Model      & \begin{tabular}[c]{@{}c@{}}Vision\\encoder\end{tabular} & \begin{tabular}[c]{@{}c@{}}Scene text\\encoder\end{tabular} & \begin{tabular}[c]{@{}c@{}}Input\\size\end{tabular} \\
\hline
ViSTA-S & $12$ layers, $6$ heads   & BERT-mini & 224$\times$ 224  \\
ViSTA-B & $12$ layers, $12$ heads   & BERT-Base & 384$\times$ 384   \\
ViSTA-L & $12$ layers, $24$ heads  & BERT-Base & 384$\times$ 384   \\
\hline
\end{tabular}}
\end{table}


\begin{table*}[thb]
\centering
\begin{threeparttable}
\caption{Comparisons with the state-of-the-art scene text aware approaches on CTC.}\vspace{-0.5em}
\label{Tab-scene-IRTR}

\footnotesize
\setlength{\tabcolsep}{1.6mm}{
\begin{tabular}{c|cccccc|cccccc} 
\hline
\multirow{3}{*}{Model} & \multicolumn{6}{c|}{CTC-1K}  & \multicolumn{6}{c}{CTC-5K} \\
& \multicolumn{3}{c}{Image-to-text} & \multicolumn{3}{c|}{Text-to-image} & \multicolumn{3}{c}{Image-to-text} & \multicolumn{3}{c}{Text-to-image}  \\
 & R@1  & R@5  & R@10   & R@1  & R@5  & R@10   & R@1  & R@5  & R@10  & R@1  & R@5  & R@10  \\ 
\hline
SCAN \cite{SCAN}  & 36.3 & 63.7 & 75.2 & 26.6 & 53.6 & 65.3 & 22.8 & 45.6 & 54.3 & 12.3 & 28.6 & 39.9    \\
VSRN \cite{VSRN}  &   38.2 & 67.4 & 79.1 & 26.6 & 54.2 & 66.2 & 23.7 & 47.6 & 59.1 & 14.9 & 34.7 & 45.5     \\
STARNet \cite{STARNet}  &44.1 &74.8& 82.7& 31.5& 60.8& 72.4& 26.4& 51.1& 63.9& 17.1& 37.4& 48.3 \\

\hline
ViSTA-S   &  \bf{52.5} & \bf{77.9} & \bf{87.2} & \bf{36.7}  &  \bf{66.2}  &  \bf{77.8}  & \bf{31.8} & \bf{56.6} & \bf{67.8} & \bf{20.0} & \bf{42.9} & \bf{54.4} \\

\hline
\end{tabular}}
\end{threeparttable}
\end{table*}

\subsection{Cross-Modal Contrastive Learning}
\label{Cross-Modal Contrastive Learning}

Conventional scene text free and scene text aware image-text retrieval are two different tasks calling for visual features only and fused visual-semantic features respectively, which correspond with the output features of [IMG] and [FUS] tokens. The final features are constructed into image-text pairs or fusion-text pairs with text queries. 
We introduce dual contrastive learning losses to embed both image-text pairs and fusion-text pairs into a common cross-modal space.
The total loss is
\begin{equation}
    \mathcal{L}_{total} = \alpha \mathcal{L}_{itc} + (1-\alpha) \mathcal{L}_{ftc},
\end{equation}
where $\mathcal{L}_{itc}$ and $\mathcal{L}_{ftc}$ are image-text contrastive loss and fusion-text contrastive loss. Note that $\alpha$ is the parameter to trade off between these losses and is set to $0.9$ as default. 

For $N$ image and text pairs as a batch, the fusion-text contrastive loss aims to maximize the similarity between $N$ matched pairs and minimize the similarity between the last $N^2-N$ incorrect pairs, formulated as\vspace{-0.5em} 
\begin{equation}
\begin{aligned}
    \mathcal{L}_{ftc} = \frac{1}{2} (\mathcal{L}_{f2t} + \mathcal{L}_{t2f}).
\end{aligned}
\end{equation}
The fusion-text contrastive learning aims to minimize the symmetric loss between the fused token and text [CLS] as\vspace{-0.5em}
\begin{equation}
\begin{aligned}
    &\mathcal{L}_{f2t} = - \frac{1}{N} \sum_{i=1}^{N} log \frac{\mathrm{exp}(f^\top_i t_i / \sigma)}{\sum_{j=1}^{N} \mathrm{exp}(f^\top_i t_j / \sigma )} \\
    &\mathcal{L}_{t2f} = - \frac{1}{N} \sum_{i=1}^{N} log \frac{\mathrm{exp}(t^\top_i f_i / \sigma)}{\sum_{j=1}^{N} \mathrm{exp}(t^\top_i f_j / \sigma )}, \\
\end{aligned}
\end{equation}
where $f_i$ and $t_j$ are the normalized embedding of fusion features in the $i$-th pairs and that of text in the $j$-th pairs, respectively. The temperature parameter $\sigma$ is a trainable variable and its initial value is set to $0.07$ as default \cite{ALIGN}. 
Same as $\mathcal{L}_{ftc}$, the image-text contrastive loss is formulated as \vspace{-0.5em}
\begin{equation}
\begin{aligned}
    \mathcal{L}_{itc} = \frac{1}{2} (\mathcal{L}_{i2t} + \mathcal{L}_{t2i}),
\end{aligned}
\end{equation}\vspace{-0.5em}
where\vspace{-1em}
\begin{equation}
\begin{aligned}
    &\mathcal{L}_{i2t} = - \frac{1}{N} \sum_{i=1}^{N} log \frac{\mathrm{exp}(v^\top_i t_i / \sigma)}{\sum_{j=1}^{N} \mathrm{exp}(v^\top_i t_j / \sigma )} \\
    &\mathcal{L}_{t2i} = - \frac{1}{N} \sum_{i=1}^{N} log \frac{\mathrm{exp}(t^\top_i v_i / \sigma)}{\sum_{j=1}^{N} \mathrm{exp}(t^\top_i v_j / \sigma )}. \\
\end{aligned}
\end{equation}
Note that $v_i$ is the normalized embedding of the $i$-th image. In the training stage, if the extracted OCR result is None, the $\mathcal{L}_{ftc}$ loss would not be added to the total loss. 


\begin{table*}[thb]
\caption{Comparisons with other approaches on Flickr30K and MSCOCO in terms of zero-shot retrieval.}\vspace{-0.5em}
\label{Tab-IRTR-zeroshot}
\centering
\footnotesize
\setlength{\tabcolsep}{1.3mm}{
\begin{tabular}{c|c|cccccc|cccccc} 
\hline
\multirow{3}{*}{Model} & \multicolumn{1}{c|}{\multirow{3}{*}{\begin{tabular}[c]{@{}c@{}}Time\\(ms)\end{tabular}}} & \multicolumn{6}{c|}{Flickr30k~(1K)}  & \multicolumn{6}{c}{MS-COCO~(5K)} \\
& \multicolumn{1}{c|}{}& \multicolumn{3}{c}{Image-to-text} & \multicolumn{3}{c|}{Text-to-image} & \multicolumn{3}{c}{Image-to-text} & \multicolumn{3}{c}{Text-to-image}  \\
& \multicolumn{1}{c|}{} & R@1  & R@5  & R@10   & R@1  & R@5  & R@10   & R@1  & R@5  & R@10  & R@1  & R@5  & R@10  \\ 
\hline
ViL-BERT \cite{vilbert}    & \textasciitilde{}900      & 31.9   & 61.1  & 72.8  & -    & -    & -     & -    & -    & -     & - & - & -   \\
Unicoder-VL \cite{unicoder}   & \textasciitilde{}925      & 64.3 & 85.8 & 92.3     & 48.4 & 76.0   & 85.2    & -    & -    & -      & -    & -    & -      \\
ImageBERT \cite{imagebert}             & \textasciitilde{}900      & 70.7 & 90.2 & 94.0     & 54.3 & 79.6 & 87.5     & 44.0   & 71.2 & 80.4   & 32.3 & 59.0   & 70.2   \\
UNITER-B \cite{UNITER}              & \textasciitilde{}900      & \bf{80.7} & \bf{95.7} & 98.0   & 66.2 & 88.4 & 92.9     & -    & -    & -     & -    & -    & -   \\
ViLT-B \cite{ViLT}             & \textasciitilde{}15       & 73.2 & 93.6 & 96.5    & 55.0   & 82.5 & 89.8    & 56.5 & 82.6 & 89.6   & 40.4 & 70.0   & 81.1    \\

\hline
ViSTA-B  & \textasciitilde17  &   75.3   & 93.8 & 97.5    & 59.5 & 84.3 & 90.3       &  60.7 & 85.8 & 92.3 & 44.8 & 72.8 & 82.5 \\
ViSTA-L & \textasciitilde{}40 & 79.2 & 95.4 & \bf{98.1} & \bf{67.0} &  \bf{88.7} & \bf{93.1} & \bf{63.9}	& \bf{87.1} & \bf{93.0} & \bf{47.4} & \bf{75.0} & \bf{84.0} \\
\hline
\end{tabular}}
\end{table*}

\begin{table*}[thb]
\caption{Comparisons with state-of-the-art approaches on fine-tuning Flicker30K and MSCOCO benchmarks.}\vspace{-0.5em}
\label{Tab-IRTR}
\centering
\footnotesize
\setlength{\tabcolsep}{1.3mm}{
\begin{tabular}{c|c|cccccc|cccccc} 
\hline
\multirow{3}{*}{Model} & \multicolumn{1}{c|}{\multirow{3}{*}{\begin{tabular}[c]{@{}c@{}}Time\\(ms)\end{tabular}}} & \multicolumn{6}{c|}{Flickr30K~(1K)}  & \multicolumn{6}{c}{MS-COCO~(5K)} \\
& \multicolumn{1}{c|}{}& \multicolumn{3}{c}{Image-to-text} & \multicolumn{3}{c|}{Text-to-image} & \multicolumn{3}{c}{Image-to-text} & \multicolumn{3}{c}{Text-to-image}  \\
& \multicolumn{1}{c|}{} & R@1  & R@5  & R@10   & R@1  & R@5  & R@10   & R@1  & R@5  & R@10  & R@1  & R@5  & R@10  \\ 
\hline
SCAN \cite{SCAN} & -  & 67.4     &90.3      &95.8    & 48.6     &    77.7  & 85.2      &  50.4    &  82.2    & 90.0     &    38.6  & 69.3     & 80.4      \\
VSRN \cite{VSRN} & -  & 71.3     &  90.6    & 96.0   &  54.7    & 81.8     &  88.2     & 53.0     &   81.1   & 89.4     &40.5      & 70.6     & 81.1      \\
IMRAM \cite{IMRAM}  & - & 74.1 & 93.0 & 96.6 & 53.9 & 79.4 & 87.2 & 53.7 & 83.2 & 91.0 & 39.7 & 69.1 & 79.8 \\
GSMN \cite{GSMN}  & - & 76.4 & 94.3 & 97.3 & 57.4 & 82.3 & 89.0  & - & - & - & - & - & - \\
SGRAF \cite{SGRAF}  & - & 77.8 & 94.1 & 97.4 & 58.5 & 83.0 & 88.8 & 57.8 & - & 91.6 & 41.9 & - & 81.3 \\
\hline
Vil-BERT \cite{vilbert} & \textasciitilde{}920 & 58.2 & 84.9 & 91.5     &   -   &   -   &     -   &    -  &    -  &    -    &   -   &   -   &    - \\
Unicoder-VL \cite{unicoder}   & \textasciitilde{}925  & 86.2 & 96.3 & 99.0 & 71.5 & 91.2  &95.2     & 62.3    & 87.1    & 92.8  & 48.4    & 76.7    & 85.9      \\
UNITER-B \cite{UNITER}   & \textasciitilde 900  & 85.9 & 97.1 & 98.8  & 72.5 & 92.4 & 96.1 & 64.4 & 87.4 & 93.1  & 50.3 & 78.5 & 87.2  \\
ERNIE-ViL-B\cite{ERNIE-ViL} &  \textasciitilde 920  &  86.7  & 97.8   & 99.0 & 74.4 & 92.7 & 95.9 &    -  &    -  &    -    &   -   &   -   &  -  \\
VSEinfty \cite{VSE/infty}  & & 88.4  &   98.3   &   99.5   &     74.2 &   93.7   &  96.8     &   66.4  &  89.3 &    -    &   51.6  &   79.3   &  -  \\
PCME \cite{PCME} & -  &   -   &   -   &     - &   -   &   -   &     -   &    44.2  &    73.8  &    83.6   &   31.9   &   62.1   &  74.5  \\
Miech et al\cite{Thinking_Fast_and_Slow}  & -  &   -   &   -   &     - &   72.1 &  91.5   &     95.2  &    -  &    -  &    -    &   -   &   -   &  -  \\

12-in-1 \cite{12-in-1}  & -  &   -   &   -   &     - &   67.9   &   89.6   &     94.2   &    - &    -  &    -    &   -   &   -   &  -  \\

Pixel-BERT-X \cite{Pixel-BERT}     & \textasciitilde{}160   &  87.0 & \bf{98.9} & 99.5     &   71.5 & 92.1 & 95.8    &  63.6 & 87.5 & 93.6      &    50.1 & 77.6 & 86.2    \\
SOHO \cite{SOHO}  & -  &  86.5 & 98.1 & 99.3  &  72.5 & 92.7 & 96.1   &  66.4 & 88.2 & 93.8  & 50.6 & 78.0 & 86.7     \\
H Xue et al.\cite{xue2021probing}  & -  &   87.0  & 98.4   &  99.5 &  73.5  &  93.1   &  96.4  &    -  &    -  &    -    &   -   &   -   &  -  \\

Pixel-BERT-R \cite{Pixel-BERT}   &  \textasciitilde{}60 & 75.7  & 94.7  & 97.1 &53.4 & 80.4 & 88.5   &  59.8    &  85.5    &91.6  &41.1     & 69.7    & 80.5      \\

ViLT-B \cite{ViLT}             & \textasciitilde{}15       & 83.5 & 96.7 & 98.6   & 64.4   & 88.7 & 93.8   & 61.5 & 86.3 & 92.7  & 42.7 & 72.9   & 83.1    \\
\hline
ViSTA-B  & \textasciitilde{}17  &  84.8 & 97.4 & 99.0 & 68.9 & 91.1 & 95.1 &    63.9 &  87.8 & 93.6 & 47.8 & 75.8 & 84.5 \\
ViSTA-L & \textasciitilde{}40 & \bf{89.5} & 98.4 & \bf{99.6} & \bf{75.8} & \bf{94.2} & \bf{96.9} & \bf{68.9} & \bf{90.1} & \bf{95.4} & \bf{52.6} & \bf{79.6} & \bf{87.6} \\
\hline
\end{tabular}}
\end{table*}

\section{Experiments}
We conduct experiments on two downstream cross-modal retrieval benchmarks to validate the effectiveness of the proposed approach. The scene text aware cross-modal retrieval task is evaluated on the COCO-Text Captioned (CTC) \cite{STARNet} dataset, and the conventional cross-modal retrieval experiments are conducted on the Flickr30K \cite{flickr30k} and MSCOCO \cite{MS-COCO-5K} benchmarks, including image-to-text and text-to-image retrieval tasks reported in Tab.~\ref{Tab-scene-IRTR} and Tab.~\ref{Tab-IRTR}. We also analyze the effectiveness of structures of the proposed ViSTA and show some 
cases in the ablation study.

\noindent\textbf{Datasets}. All the pre-training, fine-tuning, and test settings of different tasks are reported in Tab.~\ref{Tab-setting}. The setting of scene text aware cross-modal retrieval task follows \cite{STARNet}. In conventional scene text free cross-modal retrieval task, four publicly available datasets including Microsoft COCO (MSCOCO) \cite{MSCOCO_dataset}, Visual Genome (VG) \cite{VG_dataset}, SBU Captions (SBU) \cite{SBU_dataset}, and Google Conceptual Captions (GCC) \cite{GCC_dataset} datasets are used for pre-training. Since the CTC dataset is also constructed from MSCOCO, all images of CTC-5K test are contained in MSCOCO train set. Therefore, for evaluation purpose on the CTC dataset, we remove the duplicate images from the MSCOCO dataset and denote it as MSCOCO$^\ast$. For the evaluation metric, all these experiments are evaluated in terms of the percentage of containing a matched pair in the top returns, i.e., $R@1$, $R@5$, and $R@10$, respectively. 

\noindent\textbf{Implementation details.} 
For a fair comparison, we implement several versions of models with different scales, as shown in Tab.~\ref{Tab-setting-model}. For all experiments, we use the AdamW optimizer with a base learning rate of $1$e-$4$ and augmentation of random horizontal flipping and random augmentation \cite{Pixel-BERT}. We pre-train for $80$ epochs on $40$ NVIDIA Tesla V100 GPUs and finetune for another $10$ epochs on $8$ Tesla V100 GPUs. For scene text free cross-modal retrieval task, we pre-train ViSTA-B and ViSTA-L on combined datasets, i.e., SBU, CC, VG and MSCOCO$^\ast$ for fair comparisons with previous approaches. Note that images in the CTC train set are all included in the train set of MSCOCO.


\subsection{Scene Text aware Cross-Modal Retrieval}
For fair comparisons in scene text aware retrieval, we evaluate models on CTC-1K and CTC-5K test sets, respectively, strictly following the previous train and test split \cite{STARNet}.
As shown in Tab.~\ref{Tab-scene-IRTR}, Our ViSTA-S model performs a large improvement of 8.4\% and 5.4\% on R@1 in scene text aware image-text retrieval task on CTC-1k. 
Compared to STARNet \cite{STARNet} which uses GCN to get the representation of scene text for fusion, we use BERT to refine it.
And the self-attention operators on vision encoders learn the long-range dependence in images and help our ViSTA model to learn the relationship between patches.
The vision and scene text aggregation layers learn the joint distribution of modalities of vision and scene text and refine the representation space.

\subsection{Scene Text Free Cross-Modal Retrieval}
For conventional image-text retrieval, we measure zero-shot and fine-tuned performance on the Karpathy \& Fei-Fei split of MS-COCO and Flickr30K \cite{flickr30k} and compare with state-of-the-art methods in Tab.~\ref{Tab-IRTR-zeroshot} and Tab.~\ref{Tab-IRTR}, respectively. All the settings are the same as the pre-training stage. When fine-tuning on Flickr30K, we use the fine-tuned weight on COCO-5K as the initial weight. Following the efficient framework of dual-tower and patch projection operator, our model has a comparable speed with ViLT \cite{ViLT} and better performance, as shown in Tab.~\ref{Tab-IRTR}. And our large-scale model ViSTA-L achieves superior results than state-of-the-art methods at a low speed. 
Our model is not affected in those datasets when the modality of scene text is missing and still performs well on downstream tasks due to the fusion token based vision and scene text aggregation.

\subsection{Ablations}
To validate the effectiveness of the proposed vision and scene text aggregation layers for the visual tower, we conduct ablation experiments on the CTC dataset. We fix the text tower with BERT-mini and implement different visions of the visual tower. 
As shown in Tab.~\ref{Tab-ab-scene-embs-fusion}, only using scene text information encoded by GCN or BERT-mini is insufficient for cross-modal retrieval. Compare with the architecture in STARNet \cite{STARNet}, incorporating vision transformer in cross-modal retrieval, e.g., ViT-S, can achieve better performance due to the improved visual representation. Compared with results of only using the visual modality, ViSTA with scene text embedding can remarkably improve the performance by $5.5\%$/$2.1\%$ in R@1 on CTC-1K, which is contributed by the effective vision and scene text aggregation.

\begin{figure*}[ht]
  \footnotesize{Text query: A tennis team was featured in a newspaper in 1970 or 1971.}\\
  \begin{subfigure}{.1585\textwidth}
    \centering
    \includegraphics[height = 2cm, width = 3cm]{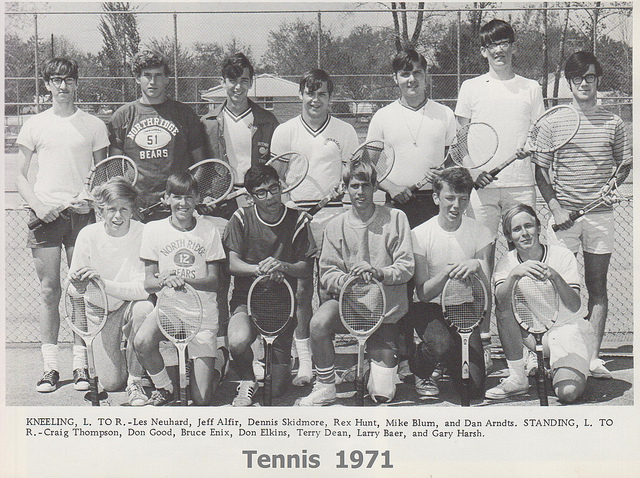}
    \caption{Top-1 \textcolor{blue}{\XSolidBrush}}
    \label{sf1}
  \end{subfigure}
  \begin{subfigure}{.1585\textwidth}
    \centering
    \includegraphics[height = 2cm, width = 3cm]{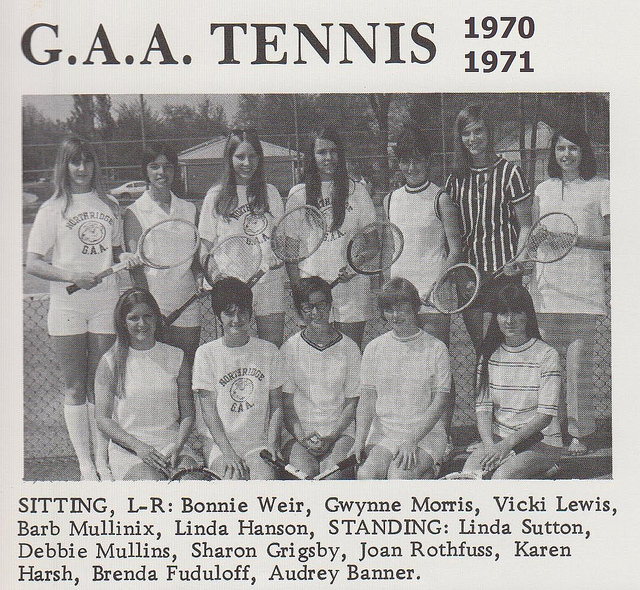}
    \caption{Top-2 \textcolor{red}{\CheckmarkBold}}
    \label{sf2}
  \end{subfigure}
  \begin{subfigure}{.1585\textwidth}
    \centering
    \includegraphics[height = 2cm, width = 2.9cm]{figures/img5_4.jpg}
    \caption{Top-3 \textcolor{blue}{\XSolidBrush}}
    \label{sf2}
  \end{subfigure}
  \hspace{2mm}
  \begin{subfigure}{.1585\textwidth}
    \centering
    \includegraphics[height = 2cm, width = 3cm]{figures/img5_1.jpg}
    \caption{Top-1 \textcolor{red}{\CheckmarkBold}}
    \label{sf1}
  \end{subfigure}
  \begin{subfigure}{.1585\textwidth}
    \centering
    \includegraphics[height = 2cm, width = 3cm]{figures/img5_2.jpg}
    \caption{Top-2 \textcolor{blue}{\XSolidBrush}}
    \label{sf2}
  \end{subfigure}
  \begin{subfigure}{.1585\textwidth}
    \centering
    \includegraphics[height = 2cm, width = 3cm]{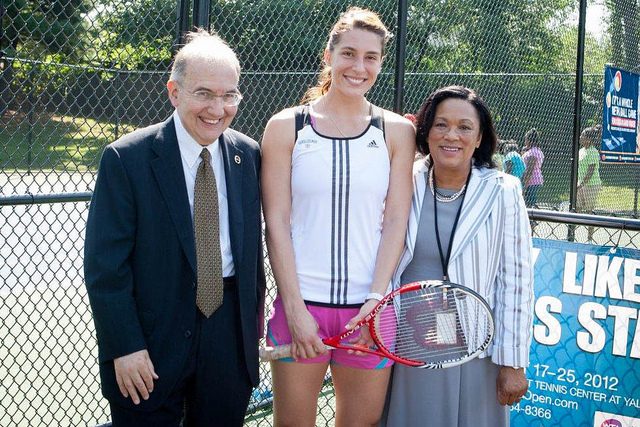}
    \caption{Top-3 \textcolor{blue}{\XSolidBrush}}
    \label{sf2}
  \end{subfigure}\vspace{0.5em}\\
\footnotesize{Text query: a person holding up a gummy hot dog in their hand}\\
  \begin{subfigure}{.1585\textwidth}
    \centering
    \includegraphics[height = 2cm, width = 3cm]{figures/img3_2.jpg}
    \caption{Top-1 \textcolor{blue}{\XSolidBrush}}
    \label{sf1}
  \end{subfigure}
  \begin{subfigure}{.1585\textwidth}
    \centering
    \includegraphics[height = 2cm, width = 3cm]{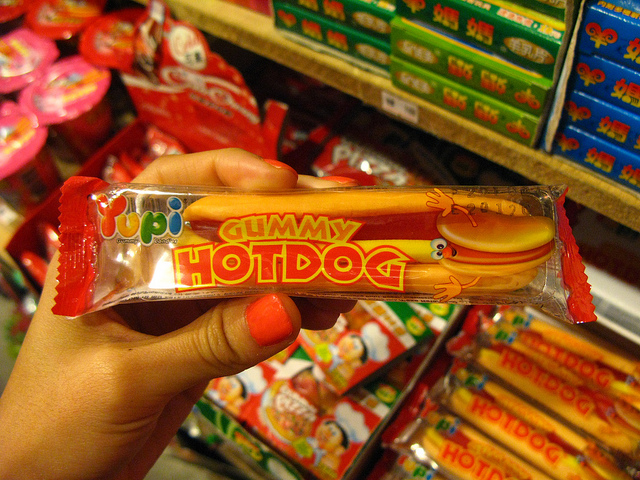}
    \caption{Top-2 \textcolor{red}{\CheckmarkBold}}
    \label{sf2}
  \end{subfigure}
  \begin{subfigure}{.1585\textwidth}
    \centering
    \includegraphics[height = 2cm, width = 3cm]{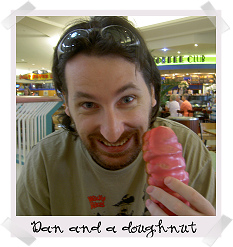}
    \caption{Top-3 \textcolor{blue}{\XSolidBrush}}
    \label{sf2}
  \end{subfigure}
  \hspace{2mm}
  \begin{subfigure}{.1585\textwidth}
    \centering
    \includegraphics[height = 2cm, width = 3cm]{figures/img3_1.jpg}
    \caption{Top-1 \textcolor{red}{\CheckmarkBold}}
    \label{sf1}
  \end{subfigure}
  \begin{subfigure}{.1585\textwidth}
    \centering
    \includegraphics[height = 2cm, width = 3cm]{figures/img3_3.jpg}
    \caption{Top-2 \textcolor{blue}{\XSolidBrush}}
    \label{sf2}
  \end{subfigure}
  \begin{subfigure}{.1585\textwidth}
    \centering
    \includegraphics[height = 2cm, width = 3cm]{figures/img3_2.jpg}
    \caption{Top-3 \textcolor{blue}{\XSolidBrush}}
    \label{sf2}
  \end{subfigure}
  \vspace{-0.5em}
  \caption{Examples of the text-to-image retrieval task for comparisons between results with and without scene text. Note that text queries with the corresponding top returned images are shown in (a) to (l). The first three columns show the retrieved results of ViSTA-S without scene text embedding, and the last three columns show the results of ViSTA-S. (best view in colors).}
  \label{fig:text2image-demo}
\end{figure*}

\begin{figure*}
  \begin{subfigure}{.32\textwidth}
    \centering
    \includegraphics[height = 3.7cm, width = 5.5cm]{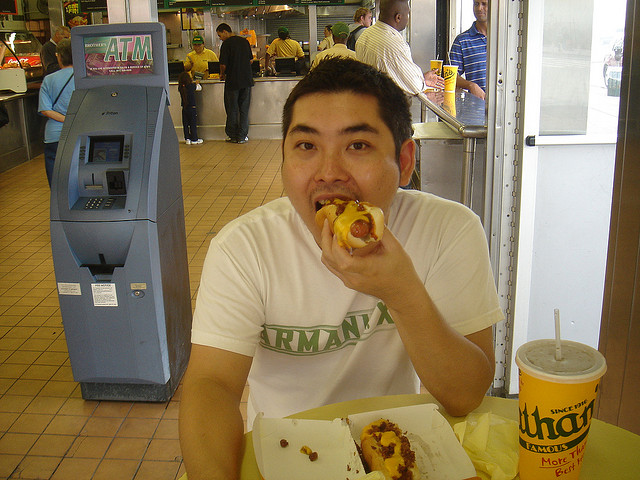}
    \caption{Top three returned results by ViSTA:\\
    \tiny
    1) A man eating a Nathans chili cheese dog in front of an \textbf{ATM}. \textcolor{red}{\CheckmarkBold} \\
    2) A man eats a hot dog at a fast food place. \textcolor{blue}{\XSolidBrush}\\
    3) The guy is eating a doughnut at a doughnut shop. \textcolor{blue}{\XSolidBrush} \vspace{5pt} \\ 
    \scriptsize
    Top three returned results of ViSTA w/o scene text:\\
    \tiny
    1) The guy is eating a doughnut at a doughnut shop. \textcolor{blue}{\XSolidBrush} \\
    2) A man eats a hot dog at a fast food place. \textcolor{blue}{\XSolidBrush}\\
    3) A young man biting a hot dog sitting at a table at a fast food court. \textcolor{red}{\CheckmarkBold}
    }
    \label{sf1}
  \end{subfigure}
  \hspace{3pt}
  \begin{subfigure}{.32\textwidth}
    \centering
    \includegraphics[height = 3.7cm, width = 5.6cm]{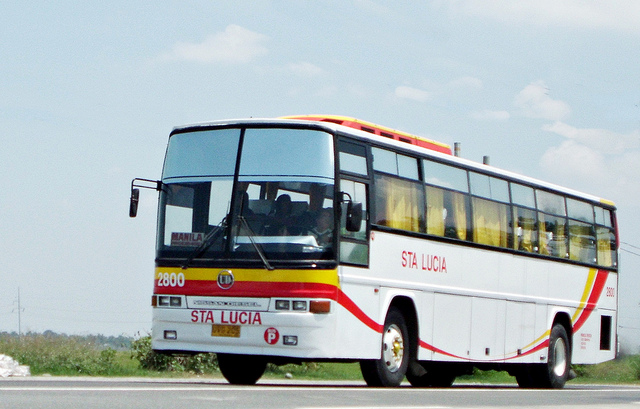}
    \caption{Top three retrieved results by ViSTA:\\
    \tiny
    1) A \textbf{STA LUCIA} bus is driving down the road. \textcolor{red}{\CheckmarkBold}\\
    2) A bus sits in the parking lot outside of Piccadilly Gardens. \textcolor{blue}{\XSolidBrush}\\
    3) A charter bus with two stories heading to some where. \textcolor{blue}{\XSolidBrush} \vspace{5pt}\\
    \scriptsize
    Top three returned results of ViSTA w/o scene text:\\
    \tiny
    1) A bus pull into a small parking lot space. \textcolor{blue}{\XSolidBrush} \\
    2) A charter bus with two stories heading to some  where.\textcolor{blue}{\XSolidBrush} \\
    3) A bus sits in the parking lot outside of Piccadilly Gardens. \textcolor{blue}{\XSolidBrush}}
    \label{sf2}
  \end{subfigure}
  \hspace{3pt}
   \begin{subfigure}{.35\textwidth}
    \centering
    \includegraphics[height = 3.7cm, width = 5.8cm]{figures/img2_4.jpg}
    \caption{Top three returned results by ViSTA:\\
     \tiny
    1) The arriving passengers on the Ethiopian airliner are deplaning on the runway. \textcolor{blue}{\XSolidBrush} \\
    2) A \textbf{China} Airlines airliner is parked at an airport near another jet. \textcolor{red}{\CheckmarkBold}\\
    3) A large continental jet sitting on a tarmac at an airport. \textcolor{blue}{\XSolidBrush} \vspace{5pt} \\ 
    \scriptsize
    Top three returned results of ViSTA w/o scene text:\\
    \tiny
    1) Commercial Lufthansa air plane parked at an airport. \textcolor{blue}{\XSolidBrush} \\
    2) The arriving passengers on the Ethiopian airliner are deplaning on the runway. \textcolor{blue}{\XSolidBrush}\\
    3) An Aegean Airlines airplane on an airport runway. \textcolor{blue}{\XSolidBrush}
    }
    \label{sf1}
  \end{subfigure}\vspace{-0.5em}
  \caption{Examples of image-to-text retrieval for comparisons between the top returned results with and without scene text.}\vspace{-0.5em}
  \label{fig:image2text-demo}
\end{figure*}

\begin{table}[thb]
\caption{Ablation study on the impact of modality aggregation.}\vspace{-0.3em}
\label{Tab-ab-scene-embs-fusion}
\centering
\footnotesize
\setlength{\tabcolsep}{0.8mm}{
\begin{tabular}{c|c|c|cccccc} 
\hline
\multicolumn{1}{c|}{\multirow{3}{*}{\begin{tabular}[c]{@{}c@{}}Model \end{tabular}}} & \multicolumn{1}{c|}{\multirow{3}{*}{\begin{tabular}[c]{@{}c@{}}Visual\end{tabular}}} & \multicolumn{1}{c|}{\multirow{3}{*}{\begin{tabular}[c]{@{}c@{}}Scene\\text \end{tabular}}} & \multicolumn{6}{c}{CTC-1K}   \\
& & \multicolumn{1}{c|}{} & \multicolumn{3}{c}{Image-to-text} & \multicolumn{3}{c}{Text-to-image} \\
& & \multicolumn{1}{c|}{} & R@1 & R@5 & R@10 & R@1  & R@5 & R@10  \\ 
\hline
GCN &    &   \checkmark   &  10.8 & 20.2 & 25.4  & 4.4 & 11.3 & 15.6  \\
BERT-mini &             &   \checkmark   & 24.3 & 35.4 & 40.8 & 9.6 & 17.8 & 22.6   \\
RoI + GCN \cite{STARNet} & \checkmark  & \checkmark   & 44.1 & 74.8 & 82.7 & 31.5 & 60.8 & 72.4  \\
ViT-S + GCN & \checkmark  &    \checkmark  &  47.2 & 74.2 & 84.2  & 33.2 & 63.6 & 75.4   \\
ViSTA-S & \checkmark  & ~ & 47.0 & 73.8 & 84.3 &  34.6 & 63.4 & 75.3     \\
ViSTA-S & \checkmark  &  \checkmark  & \bf{52.5} & \bf{77.9} & \bf{87.2} & \bf{36.7} & \bf{66.2} & \bf{77.8} \\
\hline
\end{tabular}}
\end{table}

\begin{table}[thb]
\caption{Ablation study on the number of fusion layers.}\vspace{-0.5em}
\label{Tab-ab-fusion-layers}
\centering
\footnotesize
\setlength{\tabcolsep}{1.2mm}{
\begin{tabular}{c|cccccc} 
\hline
 \multicolumn{1}{c|}{\multirow{3}{*}{\begin{tabular}[c]{@{}c@{}}The number\\of fused layers \end{tabular}}} & \multicolumn{6}{c}{CTC-1K}   \\
 \multicolumn{1}{c|}{} & \multicolumn{3}{c}{Image-to-text}  &\multicolumn{3}{c}{Text-to-image} \\
\multicolumn{1}{c|}{} & R@1 & R@5 & R@10 & R@1  & R@5 & R@10     \\ 
\hline
$L_f$ = 1  & 48.2 & 74.3 &  85.0 & 35.6 & 64.8 & 76.8 \\
$L_f$ = 2 & 52.2 & 77.0 & 86.3 & 35.4 & 64.8 &76.2 \\
$L_f$ = 4  & \bf{52.5} & \bf{77.9} & \bf{87.2} & \bf{36.7} &  \bf{66.2}  &  \bf{77.8}  \\
\hline
\end{tabular}}
\end{table}

\begin{table}[thb]
\caption{Ablation study on the impact of various fusion strategies.}\vspace{-0.5em}
\label{Tab-ab-fusion-style}
\centering
\footnotesize
\setlength{\tabcolsep}{1.2mm}{
\begin{tabular}{c|cccccc} 
\hline
 \multicolumn{1}{c|}{\multirow{3}{*}{\begin{tabular}[c]{@{}c@{}}Fusion \\ strategies \end{tabular}}} & \multicolumn{6}{c}{CTC-1K}   \\
 \multicolumn{1}{c|}{} & \multicolumn{3}{c}{Image-to-text}  &\multicolumn{3}{c}{Text-to-image} \\
\multicolumn{1}{c|}{} & R@1 & R@5 & R@10 & R@1  & R@5 & R@10     \\ 
\hline
Global attention & 48.4 & 75.5 & 86.5 & 34.7 & 64.3 & 76.2 \\
Cross attention  & 50.5 & 74.4 & 84.1 & 31.1 & 59.8 & 72.9 \\
Fusion token  &  \bf{52.5} & \bf{77.9} & \bf{87.2} & \bf{36.7}  &  \bf{66.2}  &  \bf{77.8}    \\
Late fusion & 49.2  & 73.4  & 85.8 & 34.9 & 65.0  & 76.7     \\
\hline
\end{tabular}}
\end{table}

\begin{table}[thb]
\caption{Ablation study on the impact of loss functions.}\vspace{-0.5em}
\label{Tab-ab-loss}
\centering
\footnotesize
\setlength{\tabcolsep}{1.2mm}{
\begin{tabular}{c|c|cccccc} 
\hline
\multicolumn{1}{c|}{\multirow{3}{*}{$\mathcal{L}_{ftc}$}}& \multicolumn{1}{c|}{\multirow{3}{*}{$\mathcal{L}_{itc}$}} & \multicolumn{6}{c}{CTC-1K}   \\
 &\multicolumn{1}{c|}{} & \multicolumn{3}{c}{Image-to-text} & \multicolumn{3}{c}{Text-to-image} \\
 &\multicolumn{1}{c|}{} & R@1 & R@5 & R@10 & R@1  & R@5 & R@10     \\ 
\hline
\checkmark & ~ & 46.6 & 71.3 & 82.4 & 30.3 & 58.7 & 71.4 \\
\checkmark &\checkmark  &\bf{52.5} & \bf{77.9} & \bf{87.2} & \bf{36.7}  &  \bf{66.2}  &  \bf{77.8}   \\
\hline
\end{tabular}}
\end{table} 

We also conduct several experiments to validate the effectiveness of the proposed architecture.
Tab.~\ref{Tab-ab-fusion-layers} shows that the model can improve the results with an increased number of fusion layers. Tab.~\ref{Tab-ab-fusion-style} shows the results between the proposed fusion token based method, multi-modal transformer with global attention~\cite{vl-betr} and cross-attention \cite{vilbert} as well as late fusion strategy for comparisons. As shown in the Tab.~\ref{Tab-ab-loss}, our proposed dual contrastive learning performs better than a single fusion based contrastive loss. Two separate contrastive losses for scene text aware scenarios help to maintain effective cross-modal features when the scene text information is noisy or missing.

\noindent\textbf{Qualitative comparisons.} For visual comparisons, we also report some examples to illustrate the effectiveness of our method. Our model benefits from the scene text information in learning visual features. As shown in Fig.~\ref{fig:text2image-demo}, based on the query of "tennis", "1970", and "1971", our ViSTA model matches the correct images while the ViTSA without scene text embedding retrieves a confusing result. And in the second example, the "gummy hotdog" is perfectly retrieved. For the text retrieval task, shown in Fig.~\ref{fig:image2text-demo}, the scene text extracted from images has semantic information and is contained in retrieved results with ViSTA while it does not well without scene text embedding.

\vspace{-0.5em}
\section{Conclusions and Discussions}
We have proposed an effective vision and scene text aggregation transformer to learn a scene text enhanced visual representation for cross-modal learning, unifying conventional and scene text aware cross-modal retrieval tasks in a single framework. To handle images where scene text does not appear, we propose a fusion token based aggregation approach, sharing relevant information only through the fusion token, and a dual contrastive learning approach to enhance the visual features as well. Experimental results show the superior performance of ViSTA on both scene text aware retrieval and scene text free retrieval methods, which demonstrates the effectiveness of the proposed framework. 

Note that the proposed approach can be also applied in other vision and language tasks when scene text is necessary as an additional modality. The contribution from scene text aggregation also depends on the percentage of images containing relevant scene text semantics and the correlation between visual appearance and scene text in a specific task. 

\noindent\textbf{Broader impacts}. Since the proposed approach can be trained with a large amount of image and text pairs collected from the web, further data analysis, balancing and cleaning should be taken in production to mitigate the negative social impacts caused by distribution bias and mislabeled data.


{\small
\normalem
\bibliographystyle{ieee_fullname}
\bibliography{egbib}
}

\end{document}